# A Study of the Fundamental Parameters of Particle Swarm Optimizers

Mauro Sebastián Innocente[1], Johann Sienz[2]

[1, 2]Swansea University, Centre for Polymer Processing Simulation and Design, C²EC
Research Centre, Swansea, SA2 8PP, Wales – United Kingdom
[1]mauroinnocente@yahoo.com.ar
[2]J.Sienz@swansea.ac.uk

Abstract
The range of applications of traditional optimization methods are limited by the features of the object variables, and of both the objective and the constraint functions. In contrast, population-based algorithms whose optimization capabilities are emergent properties—such as evolutionary algorithms and particle swarm optimization—present almost no restriction on those features, and can handle different optimization problems with few or no adaptations. Their main drawbacks consist of their comparatively higher computational cost and difficulty in handling equality constraints. The particle swarm optimization method is sometimes viewed as an evolutionary algorithm because of their many similarities, despite not being inspired by the same metaphor: they evolve a population of individuals taking into account previous experiences and using stochastic operators to introduce new responses. The advantages of evolutionary algorithms with respect to traditional methods have been greatly discussed in the literature for decades. While the particle swarm optimizers share such advantages, their main desirable features when compared to evolutionary algorithms are their lower computational cost and easier implementation, involving no operator design and few parameters to be tuned. However, even slight modifications of these parameters greatly influence the dynamics of the swarm. This paper deals with the effect of the settings of the parameters of the particles' velocity update equation on the behaviour of the system.

**1. Introduction**
Optimization is the process of seeking the values of the variables that lead to the best performance of a model, where the valid values of the variables must usually comply with a number of constraints. Thus, setting different combinations of values of the "variables" allows trying different candidate solutions, the "constraints" limit the valid combinations, and the "optimality criterion" allows differentiating better from worse. The suitability of traditional methods is limited by the nature of the variables, and by a number of requirements that the objective and the constraint functions must comply with. Evolutionary algorithms (EAs) comprise numerous techniques developed along the last few decades, which are inspired by evolution processes that natural organisms undergo to adapt to a dynamic environment in order to survive. Since these organisms adapt by seeking the best response to the challenge they are facing, they happen to perform complex optimization processes that can be viewed as processes of fitness maximization. Although these methods typically require higher computational resources than traditional methods, they do not impose restrictions on the features of the objective and constraint functions. Besides, they are not problem-specific but general-purpose methods, which require few or no adaptations to handle different problems. In fact, they can be used to deal with both continuous and discrete problems.

On the one hand, EAs can be viewed as "modern heuristic techniques" because they are not developed in a deterministic fashion. That is to say, they are not designed to optimize a given problem but to perform some procedures which are not directly related to the optimization process. Optimization occurs, nevertheless, despite there not being clear, evident links between the implemented technique and the resulting optimization process. This leads their detractors to argue that using them implies giving up on understanding the real problem. On the other hand, they can also be viewed as "Artificial Intelligence (AI) techniques" because their ability to optimize is an emergent property that is not specifically intended, and therefore not implemented in the code. EAs are not designed to optimize but to carry out some kind of artificial evolution performing biological-like evolution processes such as mutation, recombination, and selection. This artificial evolution results in the maximization of a fitness function that resembles biological evolution.

Swarm intelligence (SI) is the branch of AI which is concerned with the study of the collective behaviour that emerges from decentralized and self-organized systems. It is the property of a system whose individual parts interact locally with one another and with their environment, inducing the emergence of coherent global patterns that the individual parts are unaware of. In other words, the latter do not have a sense of purpose of the global emergent behaviour exhibited by the whole system. The key issue is the concept of emergence, where an "emergent property" is a feature of a swarm of simple entities as a whole, which does not exist at the individual level. The interactions among a number of entities might give birth to an emergent property, which is not possible to be inferred by analyzing an isolated individual. Likewise, when designing artificial entities that would display emergent properties, such properties cannot be deterministically implemented. It is extremely difficult even to predict whether a property would emerge from certain kinds of interactions among certain kinds of entities (not to mention which property) because the interactions, which are executed based on purely local information, must generate a positive feed-back effect. Typically, a lower threshold for the number of entities involved is required for the feed-back to take place. However, the interactions may just cancel each other out.

The particle swarm optimization (PSO) method is sometimes considered as yet another EA despite not being inspired by natural evolution because it also evolves a population of individuals by profiting from previous experiences, and uses stochastic operators to introduce new responses. However, since the paradigm adheres to the principles of SI articulated by Millonas (quoted in [1]), the method is also viewed as one of the most successful SI-based optimization methods.

The PSO paradigm was originally designed by social-psychologist James Kennedy and electrical-engineer Russell Eberhart in 1995 [1]. Although it was inspired by previous bird flock simulations, the latter were framed within the field of social psychology, under the view of mind as social phenomena. Therefore, the paradigm is also closely related to other simulations of social processes, having strong roots in both AI and social psychology. From the optimization point of view, it is a global method capable of dealing with optimization problems whose solutions can be represented as points in an $n$-dimensional space. In its original version, the design variables must be real-valued, although binary and other discrete versions of the method have been also developed (e.g. [2, 3, 4, 5, 6, 7]).

The PSO approach and the EAs are population-based methods that rely on stochastic operators to introduce creativity. They are bottom-up approaches in the sense that the system's intelligent behaviour emerges in a higher level than the individuals', evolving intelligent solutions without using programmers' expertise on the subject matter. While this feature makes it difficult to understand the way optimization is actually performed, these algorithms show astonishing robustness in dealing with many kinds of complex problems that they were not specifically designed for. However, they present the disadvantage that their theoretical bases are extremely difficult to understand in a deterministic manner. Although much theoretical work has been carried out, only problem-specific and partial conclusions have been achieved in such important matters as the convergence and the tuning of the algorithms' parameters (e.g. refer to [8, 9, 10]). Refer to Innocente et al. [11] for a very short review of the PSO paradigm, while the following books are highly recommended for a general and comprehensive review: Kennedy et al. [12], Engelbrecht [13] and Clerc [14].

This paper intends to analyze the influence of the parameters of the particles' velocity update equation of the basic PSO algorithm on the overall behaviour of the swarm. This is carried out taking into consideration the evolutions of the best solution found so far and of the average among the current population of candidate solutions; the particles' ability to—and speed of—clustering; and the optimizer's reluctance to getting trapped in suboptimal solutions. The behaviour of the system heavily depends on the tuning of these parameters. For instance, the particles are more "self-confident" when assigned individuality greater than sociality, which results in greater reluctance to becoming a "follower". In turn, the latter might result in greater exploration in detriment of the speed of clustering.

## 2. Basic Algorithm

The PSO method was originated on the simulation of a simplified social milieu, where individuals (i.e. *particles*) were thought of as collision-proof birds. The population is usually referred to as the *swarm*, while the function to be optimized is called *conflict* function here due to the social-psychology metaphor that inspired the method: each individual searches the space of beliefs, seeking the minimization of the conflicts among the beliefs it holds by using the information gathered by both its own experience and those of others. Individuals indirectly seek agreement by clustering in the space of beliefs, which is—broadly speaking—the result of all the individuals imitating the most successful ones, thus becoming more similar to one another as the search progresses. The clustering is delayed by their own previous successful experiences, which the individuals are reluctant to disregard, resulting in further exploration of different combinations of beliefs.

To summarize, while the emergent properties of the PSO algorithm result from local interactions among the particles within the swarm, the behaviour of every particle can be summarized in terms of three principles:

- *Evaluate*: A particle evaluates its position in the environment, which is given by the associated value of the conflict function. Keeping the social psychology metaphor, this would stand for the conflict among its current set of beliefs.
- *Compare*: Once the particle's position in the environment is evaluated, it is not straightforward to tell how good it is. Experiments and theories in social psychology suggest that humans judge themselves by comparing to others. In other words, they tell better from worse rather than good from bad. Therefore, the particle compares the conflict among its current set of beliefs to those of other particles in its neighbourhood.
- *Imitate*: After the particle compares its conflict to its neighbours', it imitates only those whose performances are superior or somehow desirable. In the basic PSO algorithm, the particle imitates only its most successful neighbour.

These three processes are implemented within the PSO paradigm with remarkable success: the only sign of individual intelligence shown by the particles is a small memory. Thus, the basic equations of the PSO method are as follows:

$$v_{ij}^{(t)} = w \cdot v_{ij}^{(t-1)} + iw \cdot U_{(0,1)} \cdot \left(pbest_{ij}^{(t-1)} - x_{ij}^{(t-1)}\right) + sw \cdot U_{(0,1)} \cdot \left(gbest_{j}^{(t-1)} - x_{ij}^{(t-1)}\right) \quad (1)$$

$$x_{ij}^{(t)} = x_{ij}^{(t-1)} + v_{ij}^{(t)} \quad (2)$$

Where:

- $x_{ij}^{(t)}$ : coordinate *j* of the position of particle *i* at time-step *t*
- $v_{ij}^{(t)}$ : component *j* of the velocity of particle *i* at time-step *t*
- $U_{(0,1)}$ : random number generated from a uniform distribution in the range [0,1], resampled anew every time it is referenced
- *w*, *iw*, *sw* : inertia, individuality, and sociality weights
- $pbest_{ij}^{(t-1)}$, $gbest_{j}^{(t-1)}$ : coordinate *j* of the best position found by particle *i* and by any particle in the swarm, respectively, up to time-step (*t*-1)

As can be observed in Eq.(1), there are three parameters in the basic PSO algorithm which rule the behaviour of the swarm: the inertia, the individuality, and the sociality weights. Both the *individuality* and the *sociality* weights are also referred to as the *learning weights*, while their aggregation is called here the *acceleration weight*.

Given a current position of a particle, its performance is **evaluated** in terms of the conflict function. The latter is directly related to the function to be optimized so that the optimal solution coincides with the minimal conflict. In order to decide upon its next position, the particle **compares** its current conflict to those associated to both, its own and its neighbours' best previous experiences (*pbest* and *gbest*). Finally, the particle **imitates** the best experience of its most successful neighbour, but without disregarding its own. That is to say, it moves towards a weighted average of both experiences.

The relative importance assigned to the individuality and the sociality weights results in the particles exhibiting either more self-confident or more conformist behaviour. The random weights allow the introduction of creativity into the system: since they are resampled anew for each time-step, for each particle, for each component, and for each term in Eq.(1)[1], the particles display uneven, zigzagging trajectories that allow better exploration of the search-space. In addition, resampling them anew for the individuality and the sociality terms—together with setting $iw = sw$—leads to the particles alternating self-confident and conformist behaviour, without any of them taking the lead for too long.

Every particle also tends to keep its current velocity, where the strength of the tendency is governed by the inertia weight. The relative importance between the inertia and the acceleration weights results in either more explorative or more exploitative dynamics of the swarm.

In the original PSO algorithm, Kennedy et al. [1] did not consider the inertia weight (i.e. $w = 1$), and suggested setting $iw = sw = 2$. However, this algorithm presented a serious unexpected problem: the particles tended to diverge rather than cluster, so that the swarm appeared to perform a so-called *explosion*. It was found that if the components of the particles' velocities were clamped, for instance as shown in Eq.(3), the explosion was controlled and the particles ended up clustering around a solution. Therefore there is a fourth parameter, the $v_{max}$ constraint, which also influences the particles' trajectories despite being external to the particles' velocity update equation:

$$\begin{aligned} \text{if} \quad & v_{ij}^{(t)} > v_{max} \Rightarrow v_{ij}^{(t)} = v_{max} \\ \text{elseif} \quad & v_{ij}^{(t)} < -v_{max} \Rightarrow v_{ij}^{(t)} = -v_{max} \end{aligned} \qquad (3)$$

Later, Shi et al. [15] proposed the incorporation of the inertia weight to the original algorithm aiming to control the explosion. This also allowed controlling the relative importance between the inertia and the acceleration weights.

## 3. Explosion

The dynamics and reasons for the explosion to occur are still not fully understood, although they were found to be related to both the relative importance given to the inertia and the acceleration weights on the one hand, and to the random weights on the other. An example of the explosion for a 1-dimensional problem is shown in Figure 1.

Clerc et al. [8] simplified the system in order to study the dynamics of the swarm from the bottom up. Hence the random weights were simply removed, and the swarm reduced to a single particle flying over a 1-dimensional space. In addition, the particle was attracted towards two stationary best previous experiences, so that it was in reality attracted towards a fixed point that resulted from the weighted average of the two stationary attractors (*pbest* and *gbest*), as shown in Eq.(4):

$$iw \cdot pbest + sw \cdot gbest = (iw + sw) \cdot p \quad \Rightarrow \quad p = \frac{iw \cdot pbest + sw \cdot gbest}{iw + sw} \qquad (4)$$

Therefore, Eq.(1) was replaced by Eq. (5):

$$v^{(t)} = v^{(t-1)} + (iw + sw) \cdot \left(p - x^{(t-1)}\right) \qquad (5)$$

---

[1] It is important to note that mistakes in this regard can be found in the literature. In fact, even Kennedy et al. [1, 12]—the inventors of the method—make some nomenclature mistakes which mislead the reader with respect to the resampling of the random weights. Their implementations, nevertheless, consider the resampling as stated here.

They proved that if $aw = iw + sw < 4$, the particle exhibits a cyclic or quasi-cyclic behaviour. Even further, they found the particular values of $aw$ for which the behaviour is cyclic. Conversely, there is no cyclic behaviour, and the particle diverges from $p$, if $aw \geq 4$. The evolution of such a particle for $aw = 4$ and $p = 0$ is shown in Figure 2. This divergence is referred to as the "deterministic explosion" within this paper.

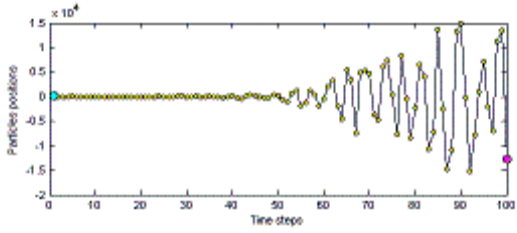 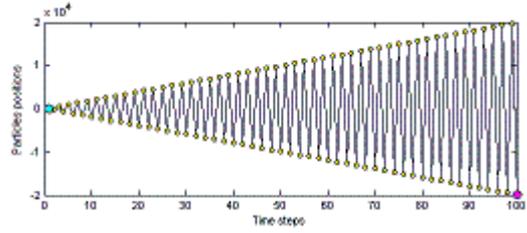

Figure 1. Evolution of a single particle flying over a 1-dimensional space, where the two best values are fixed to zero, the particle is initially located at $x = 100$, its velocity is randomly initialized within the interval $[-1,1]$, no $v_{\max}$ is imposed, $iw = sw = 2$, and the function to be optimized is the Schaffer f6. The cyan and magenta dots are the particle's initial and final positions.

Figure 2. Evolution of a single particle flying over a 1-dimensional space, where the two best values are fixed to zero, the particle is initially located at $x = 100$, its velocity is initialized to zero, no $v_{\max}$ is imposed, the random weights are removed, $iw = sw = 2$, and the function to be optimized is the Schaffer f6. The cyan and magenta dots are the particle's initial and final positions.

Clerc et al. [8] analytically developed a constriction factor that is claimed to ensure the convergence on local optima of the single non-random particle, generalizing their analytic findings to the full multi-particle system, with the random weights, and with the two non-stationary best previous experiences. These generalized algorithms were successfully tested on a set of benchmark functions. Some other researchers have also studied the trajectory of a single non-random particle (e.g. Kennedy et al. [12], Ozcan et al. [9], and Trelea et al. [10]).

Although both the explosion observed in Figure 1 and the one observed in Figure 2 occur for $aw = 4$, the latter is a purely deterministic explosion. While Clerc et al. [8] dealt with the mathematical reasons for this deterministic explosion, the dynamics of the swarm once the random weights $U_{(0,1)}$ are incorporated are not strictly considered.

The average behaviour of the PSO method according to Eq.(1) and Eq.(2), for $w = 1$, $iw = sw = 2$, and replacing the random weights by the mean of the uniform distribution used to generate them, is cyclic. This can be clearly observed in Figure 3 and Figure 4:

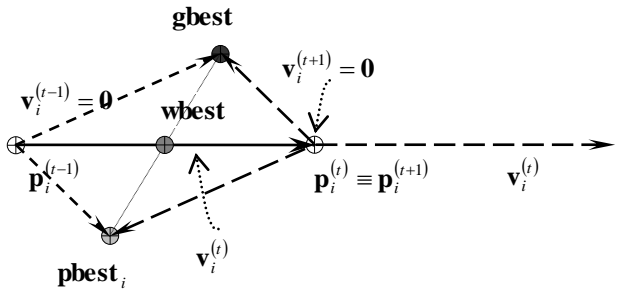 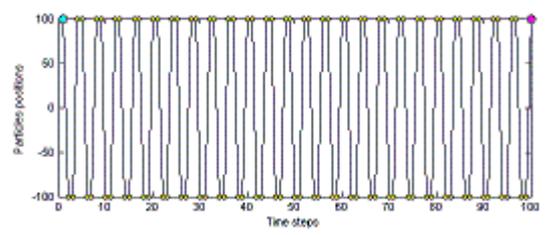

Figure 3. Sketch of the trajectory of a particle $i$, which is attracted towards the points **gbest** and **pbest**$_i$, where $iw = sw = 2$, $U_{(0,1)}$ in the velocities' update rule are replaced by $\overline{U}_{(0,1)} = 0.5$, and $\mathbf{v}_i^{(t-1)} = \mathbf{0}$. Therefore, this trajectory is in reality the part of the complete trajectory of a generic particle that is induced by the attractors at time-step $(t-1)$ (i.e. the inertia at $t-1$ is missing).

Figure 4. Evolution of a single particle flying over a 1-dimensional space, where the two best values are fixed to zero, the particle is initially located at $x = 100$, its velocity is initialized to zero, no $v_{\max}$ is imposed, $U_{(0,1)}$ is replaced by $\overline{U}_{(0,1)} = 0.5$, $iw = sw = 2$, and the function to be optimized is the Schaffer f6. The cyan and magenta dots are the particle's initial and final positions.

It is not clear why by simply incorporating the random weights instead of the constant "0.5", the particle ends up diverging rather than exhibiting a cyclic average behaviour. A simplistic heuristics argues that since each random weight generated is as likely to be greater as it is to be less than 0.5 and there is more space to explode to than to implode to, the particle is more likely to diverge. This explosion is referred to as the "probabilistic explosion" within this paper.

In order to visualize the probabilistic explosion, Figure 5 shows the trajectory of a particle with $aw = 1$ and with the random weights replaced by the mean of the distribution from which they are generated, and Figure 6 shows the trajectory of the same particle with the random weights. Note that the deterministic explosion does not occur because $aw < 4$.

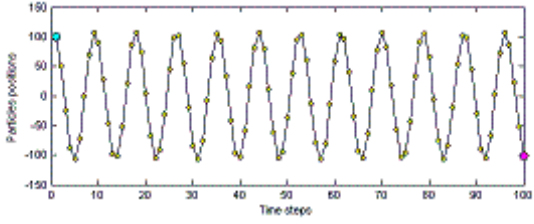

Figure 5. Evolution of a single particle flying over a 1-dimensional space, where the two best values are fixed to zero, the particle is initially located at $x = 100$, its velocity is initialized to zero, no $v_{max}$ is imposed, $U_{(0,1)}$ is replaced by $\overline{U}_{(0,1)} = 0.5$, $iw = sw = 0.5$, and the function to be optimized is the Schaffer f6.

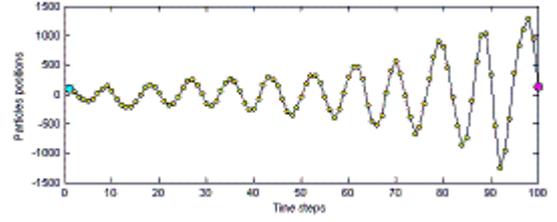

Figure 6. Evolution of a single particle flying over a 1-dimensional space, where the two best values are fixed to zero, the particle is initially located at $x = 100$, its velocity is initialized to zero, no $v_{max}$ is imposed, $iw = sw = 0.5$, and the function to be optimized is the Schaffer f6.

As previously mentioned, bracketing the components of the particles' velocities according to equation (3) effectively controls the explosion. Instead, Clerc et al. [8] proposed the incorporation of a so-called constriction factor ($\chi$) to the original version of the algorithm, claiming that it would ensure convergence. Hence their update equations are as follows:

$$\chi = \begin{cases} \dfrac{2 \cdot \kappa}{\left| (iw + sw) - 2 + \sqrt{(iw + sw)^2 - 4 \cdot (iw + sw)} \right|} & \text{if } (iw + sw) \geq 4 \\ \sqrt{\kappa} & \text{otherwise} \end{cases} \quad (6)$$

$$v_{ij}^{(t)} = \chi \cdot \left( v_{ij}^{(t-1)} + iw \cdot U_{(0,1)} \cdot \left( pbest_{ij}^{(t-1)} - x_{ij}^{(t-1)} \right) + iw \cdot U_{(0,1)} \cdot \left( gbest_{j}^{(t-1)} - x_{ij}^{(t-1)} \right) \right) \quad (7)$$

$$x_{ij}^{(t)} = x_{ij}^{(t-1)} + v_{ij}^{(t)} \quad (8)$$

Where $\chi$ is the constriction factor, and $0 < \kappa \leq 1$.

Given that both the original version of the algorithm and the one with the constriction factor can be viewed as particular cases of the version with the inertia weight, the latter is considered here as the basic PSO. Note that the parameters of the velocity update equation are not necessarily constant along the search or the same for every object variable.

The current paper is only concerned with the influence of the settings of the parameters of the particles' velocity update equation on the behaviour of the swarm. The study of other important settings and aspects of the algorithm such as the population size, the initialization procedure, the size of the neighbourhoods, the topological social structure of the swarm, the either sequential or parallel update of the swarm's best experience, the constraint-handling technique, and the termination conditions are beyond the scope of this paper.

In agreement with the swarm sizes suggested by Kennedy et al. [12] and Carlisle et al. [17], a swarm of 30 particles is used for the experiments in this article. Other features of the optimizers used here are as follows: global version; random initialization; update of the swarm's best experience only after every particle's best experience is updated (parallel); additive penalization; and the termination conditions developed in [18].

## 4. Influence of different parameters' settings

### 4.1. $v_{max}$ constraint

Although the constriction factor and the inertia weight are effective either in preventing the particles from exploding or at least in eventually pulling them back so that they end up clustering, it is widely agreed in the literature that the $v_{max}$ constraint should be kept because it prevents subsequent evaluations of the conflict function far from the region of interest. Several settings for the $v_{max}$ constraint were tried. Small values enhanced the fine-tuning of the search, while large values favoured exploration. However, the small values put at risk the ability of the optimizer to escape local optima, while large values resulted in the lack of fine-tuning the search. Linearly time-decreasing values of $v_{max}$ were effective in enhancing the accuracy of the solutions for the original version of the PSO algorithm, but did not appear to lead to much improvement once the inertia weight was incorporated. A setting frequently found in the literature (e.g. [16]) is $v_{max} = 0.5 \cdot (x_{max} - x_{min})$, which is large enough not to risk the explorative behaviour yet small enough to avoid numerous, unnecessary evaluations of the objective function. Thus, the enhancement of the fine-tuning of the search is left for the inertia, the individuality and the sociality weights.

The beneficial effect of the inertia weight for the control of the explosion can be observed in Figure 7, where the absence of the $v_{max}$ allows the explosion to take place. However, the inertia weight $w < 1$ pulls the particles back to the region of interest, and they end up clustering around a solution. The beneficial effect of the inertia weight for the fine-tuning of the search can be visualized by comparing Figure 8 and Figure 9.

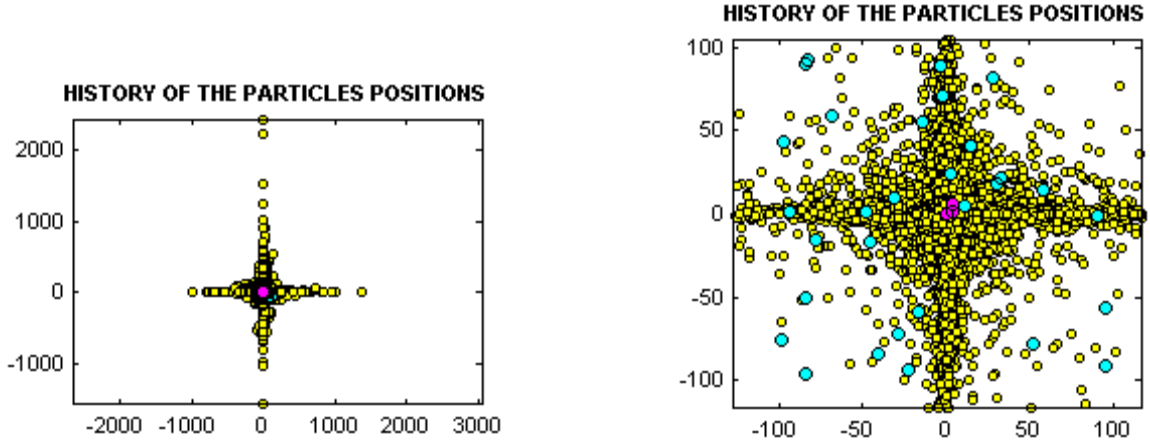

Figure 7. History of the particles' positions for $w = 0.7$, $iw = sw = 2$, and no $v_{max}$.

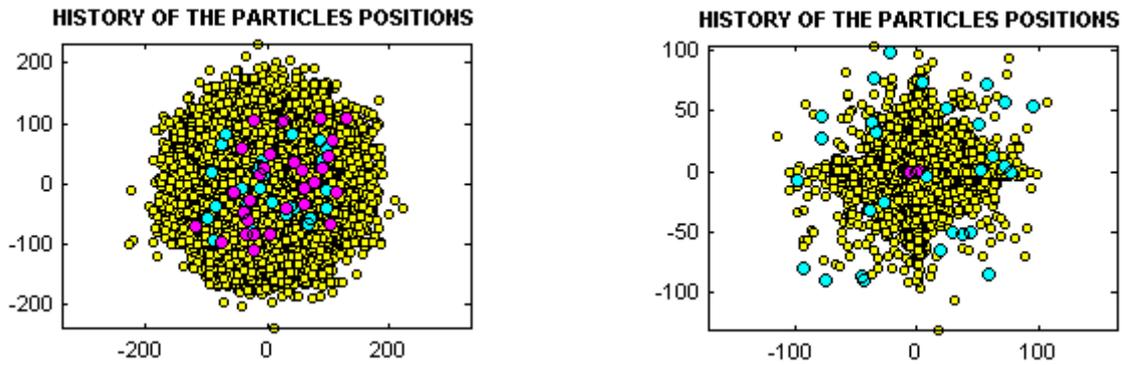

Figure 8. History of the particles' positions for $w = 1$, $iw = sw = 2$, and $v_{max} = 100$ after 4000 time-steps, when optimizing the Schaffer f6 function. The cyan and magenta dots are the initial and final particles' positions, respectively. This is equivalent to removing $w$.

Figure 9. History of the particles' positions for $w = 0.7$, $iw = sw = 2$, and $v_{max} = 100$ after 4000 time-steps, when optimizing the Schaffer f6 function. The cyan and magenta dots are the initial and final particles' positions, respectively.

4.2. Inertia weight

Several settings for the inertia weight were implemented, keeping $iw = sw = 2$ as initially proposed by Kennedy et al. [1]. While it is reasonable to favour exploration at the early stages of the search switching to a more exploitative behaviour as the search progresses, it is typically claimed in the literature that higher values of the inertia weight favour exploration and lower values favour exploitation. Hence linearly time-decreasing inertia weights (with $w^{(t)} < 1$ $\forall t$) were implemented. As expected, this led to better fine-clustering of the particles—and thus better fine-tuning of the search—compared to the fine-clustering exhibited by the particles of optimizers with high, constant inertia weights (e.g. $1 \leq w \leq 0.7$).

However, a geometrical analysis of the update of a particle's position analogous to that of Figure 3 ($w = 1$, $iw = sw = 2$) shows that the average behaviour of the swarm would be cyclic again when $w = 0$, as can be observed in Figure 10. Geometrically looking for the value of the inertia weight that would favour fast clustering, it was found out that such value would be equal to $w = 0.5$, as shown in Figure 11. Note that these analyses are valid only for $aw = 4$!

Numerous experiments were carried out optimizing the set of benchmark functions shown in Table 1, showing that the setting $w = 0.5$, $iw = sw = 2$ effectively favours fine-clustering.

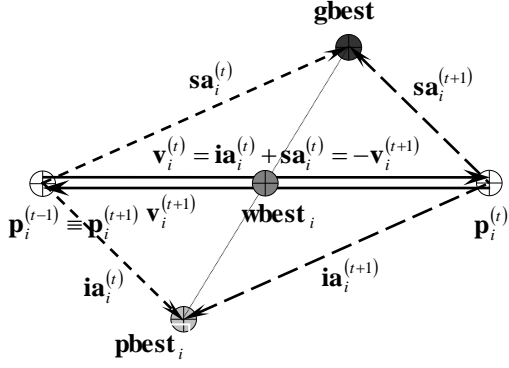
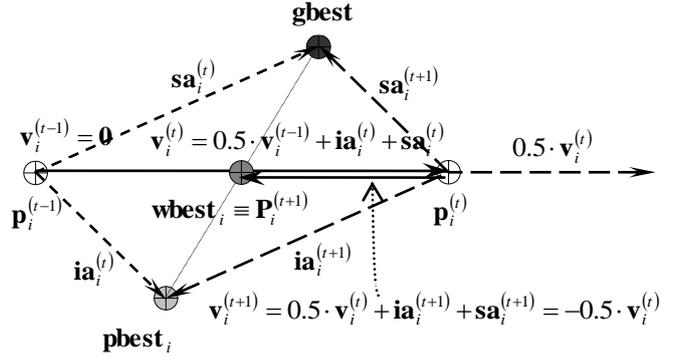

Figure 10. Sketch of the trajectory of a particle $i$, which is attracted towards the points **gbest** and **pbest**$_i$, corresponding to an optimizer with $w=0$, $iw=sw=2$, and replacing the random weights $U_{(0,1)}$ in the velocities' update equation by the average $\overline{U}_{(0,1)}=0.5$.

Figure 11. Sketch of the trajectory of a particle $i$, which is attracted towards the points **gbest** and **pbest**$_i$, corresponding to an optimizer with $w=0.5$, $iw=sw=2$, replacing the random weights $U_{(0,1)}$ in the velocities' update equation by the average $\overline{U}_{(0,1)}=0.5$, and forcing $\mathbf{v}_i^{(t-1)}=\mathbf{0}$. Therefore, this trajectory is in reality the part of the complete trajectory of a generic particle that is induced by the attractors at time-step $(t-1)$.

Table 1. Benchmark functions in the test suite

| | | | |
|---|---|---|---|
| **Sphere** | $f(\mathbf{x})=\sum_{i=1}^{n} x_i^2$ | - Search-space:<br>- Acceptable error: | $[-100,100]^{30}$<br>< 0.01 |
| **Rosenbrock** | $f(\mathbf{x})=\sum_{i=1}^{n-1} 100\cdot(x_{i+1}-x_i^2)^2+(x_i-1)^2$ | - Search-space:<br>- Acceptable error: | $[-30,30]^{30}$<br>< 100 |
| **Rastrigin** | $f(\mathbf{x})=\sum_{i=1}^{n}\left[x_i^2-10\cdot\cos(2\cdot\pi\cdot x_i)+10\right]$ | - Search-space:<br>- Acceptable error: | $[-5.12,5.12]^{30}$<br>< 100 |
| **Griewank** | $f(\mathbf{x})=\frac{1}{4000}\cdot\sum_{i=1}^{n} x_i^2-\prod_{i=1}^{n}\cos\left(\frac{x_i}{\sqrt{i}}\right)+1$ | - Search-space:<br>- Acceptable error: | $[-600,600]^{30}$<br>< 0.1 |
| **Schaffer f6** | $f(\mathbf{x})=\dfrac{\left[\sin\left(\sqrt{\sum_{i=1}^{n} x_i^2}\right)\right]^2-0.5}{\left(1+0.001\cdot\sum_{i=1}^{n} x_i^2\right)^2}+0.5$ | - Search-space:<br>- Acceptable error: | $[-100,100]^{30}$<br>< 0.1 |

While the abilities to fine-tune the search and to escape local optima are both highly desirable features, they are two sides of the same coin: optimizers that exhibit high fine-clustering ability are typically prone to get trapped in sub-optimal solutions, and optimizers with the ability to escape sub-optimal solutions usually cannot refine the search. It was found in our experiments that the setting $w=0.7$, $iw=sw=2$ results in the optimizer being able to escape sub-optimal solutions.

4.3. Learning weights

It was also proposed to linearly time-swap the relative importance between *iw* and *sw*, keeping the *aw* constant, so that the particles could exhibit higher individuality at the beginning of the search, switching to higher sociality as the search progresses. However, the results were sometimes beneficial and sometimes harmful, turning the convenience of the strategy into problem-dependent. It seems that a strong individuality might decrease the explorative behaviour of the particles for some functions. For instance, when a particle updates its best previous experience at almost every time-step and the sociality is too small, it practically moves following a straight line, nullifying its explorative ability. Therefore, a too high *iw* not always results in more explorative behaviour as it is frequently claimed in the literature. As to the *sw*, it is evident that a higher value leads to a more local search in detriment of exploration. However, it appears that too high *aw* result in reinforcing the influence of the random weights, and the dynamics of the swarm becomes more erratic.

## 4.4. Correlation between the inertia and acceleration weights

Another optimizer with strong fine-clustering ability resulted from setting $\chi = 0.7298$, $iw = sw = 2.05$ in Eq.(7), which is equivalent to setting $w = 0.7298$, $iw = sw = 1.49609$ in Eq.(1). Note that this keeps the constant correlation $w = 4.1 \cdot aw$. Aiming to find another correlation that would favour fine-clustering, 5 geometrical analyses similar to that of Figure 11 were carried out. Thus, a value of the inertia weight was derived for each acceleration weight, so that $\mathbf{p}_i^{(t+1)} \equiv \mathbf{wbest}$. An analytical function was obtained by a polynomial interpolation of the resulting pairs "$aw$-$w$":

$$aw^{(t)} = -4.142 \cdot \left(w^{(t)}\right)^4 + 12.398 \cdot \left(w^{(t)}\right)^3 - 12.77 \cdot \left(w^{(t)}\right)^2 + 7.803 \cdot w^{(t)} + 2 = p\left(w^{(t)}\right) \qquad (9)$$

Numerous experiments were carried out optimizing the functions in Table 1 for both constant and time-decreasing inertia weights in Eq.(1) ($aw = 4$); for constant and time-decreasing constriction factors in Eq.(7) ($w = 1$, $aw = 4.1$); and for constant and time-decreasing inertia weights in Eq.(1), but keeping the polynomial relationship shown in Eq.(9). For the complete set of results, refer to [18]. Due to space constraints, it is merely mentioned here that the setting $w = 0.7$, $iw = sw = 2$ results in a robust optimizer which exhibits some degree of clustering and high reluctance to getting trapped in sub-optimal solutions, while the settings $w = 0.7298$, $iw = sw = 1.49609$ and $w = 0.5$, $iw = sw = 2$ result in optimizers whose particles carry out a thorough fine-tuning of the search.

A good compromise is given by setting $iw = sw = 2$, while the inertia weight varies as shown in Figure 12:

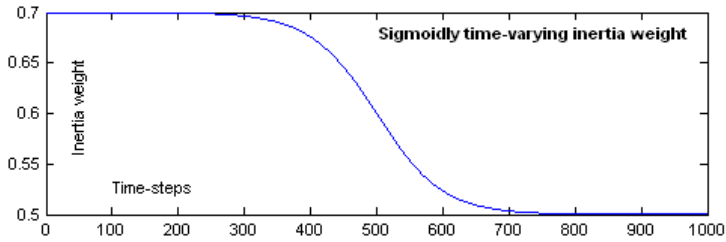

$$w(t) = \frac{(w_{\max} - w_{\min})}{1 + e^{\left(2 \cdot k \cdot \frac{t}{t_{\max}} - k\right)}} + w_{\min}$$

Figure 12. Sigmoidly time-decreasing inertia weight, where $w_{\max} = 0.7$, $w_{\min} = 0.5$ and $k = 10$.

Aiming to profit from different desirable features of different settings, it is immediate to think of a multi-swarm PSO, where the main swarm is composed of a few sub-swarms with different parameters' settings. Thus, two optimizers were implemented splitting the main swarm in three sub-swarms of the same size:

Multi-swarm 1: $w = 0.7$, $iw = sw = 2$  $\quad + \quad$ $w = 0.7298$, $iw = sw = 1.49609$ $\quad + \quad$ $w = 0.5$, $iw = sw = 2$
Multi-swarm 2: $w$ as shown in Figure 12  $\quad + \quad$ $w = 0.7298$, $iw = sw = 1.49609$ $\quad + \quad$ $w = 0.5$, $iw = sw = 2$

## 5. Experimental results

Four homogeneous and two multi-swarm optimizers are proposed and tested on the set of benchmark functions shown in Table 1. The Multi-swarm 1 optimizer is composed of the first three homogeneous optimizers, while the Multi-swarm 2 optimizer is composed of the last three ones (see previous section). The length of the search is limited to 10000 time-steps. The results are gathered in Table 2 to Table 4:

Table 2. Summary of the most significant results obtained from the optimization of the 30-dimensional Sphere and Rosenbrock functions using four selected parameters' settings and two multi-swarm optimizers (statistics out of 20 runs)

| | SPHERE | | | ROSENBROCK | | |
|---|---|---|---|---|---|---|
| OPTIMIZER | Best solution | Mean best solution | Mean time-steps | Best solution | Mean best solution | Mean time-steps |
| | Worst solution | (Standard deviation) | (Standard deviation) | Worst solution | (Standard deviation) | (Standard deviation) |
| $w = 0.7$, $iw = sw = 2$ | 2.45E-11 | 2.12E-08 | 10000.00 | 8.35E+00 | 1.10E+02 | 10000.00 |
| | 3.17E-07 | (7.03E-08) | (0.00) | 4.89E+02 | (1.20E+02) | (0.00) |
| $w = 0.7298$, $iw = sw = 1.49609$ | 1.83E-16 | 5.80E-13 | 1000.00 | 1.62E-04 | 1.82E+00 | 10000.00 |
| | 5.10E-12 | (1.26E-12) | (0.00) | 9.24E+00 | (2.35E+00) | (0.00) |
| $w = 0.5$, $iw = sw = 2$ | 1.48E-13 | 2.38E-12 | 1151.70 | 6.52E-01 | 1.61E+01 | 10000.00 |
| | 5.40E-12 | (1.47E-12) | (42.32) | 7.46E+01 | (2.02E+01) | (0.00) |
| $w \to 0.7$ to $w \to 0.5$, $iw = sw = 2$ | 1.53E-13 | 1.03E-12 | 5600.05 | 4.19E+00 | 2.63E+01 | 10000.00 |
| | 2.73E-12 | (7.40E-13) | (59.73) | 7.62E+01 | (2.54E+01) | (0.00) |
| Multi-swarm 1 (1, 2, 3) | 7.27E-65 | 1.16E-43 | 4282.20 | 1.72E-01 | 2.28E+01 | 10000.00 |
| | 1.83E-42 | (4.17E-43) | (475.67) | 8.84E+01 | (2.66E+01) | (0.00) |
| Multi-swarm 2 (2, 3, 4) | 6.65E-55 | 7.02E-40 | 3909.65 | 2.97E-03 | 1.42E+01 | 10000.00 |
| | 1.30E-38 | (2.91E-39) | (330.13) | 8.17E+01 | (2.35E+01) | (0.00) |

Table 3. Summary of the most significant results obtained from the optimization of the 30-dimensional Rastrigrin and Griewank functions using four selected parameters' settings and two multi-swarm optimizers (statistics out of 20 runs)

| OPTIMIZER | RASTRIGRIN | | | GRIEWANK | | |
|---|---|---|---|---|---|---|
| | Best solution | Mean best solution | Mean time-steps | Best solution | Mean best solution | Mean time-steps |
| | Worst solution | (Standard deviation) | (Standard deviation) | Worst solution | (Standard deviation) | (Standard deviation) |
| $w = 0.7$, $iw = sw = 2$ | 7.35E+00 | 2.18E+01 | 10000.00 | 6.00E-11 | 1.57E-02 | 10000.00 |
| | 3.78E+01 | (7.58E+00) | (0.00) | 4.67E-02 | (1.54E-02) | (0.00) |
| $w = 0.7298$, $iw = sw = 1.49609$ | 3.58E+01 | 6.04E+01 | 1048.45 | 6.79E-14 | 6.59E-02 | 1008.50 |
| | 1.11E+02 | (1.56E+01) | (107.15) | 4.05E-01 | (1.07E-01) | (26.55) |
| $w = 0.5$, $iw = sw = 2$ | 2.69E+01 | 4.94E+01 | 1760.70 | 3.47E-13 | 7.52E-03 | 1163.05 |
| | 6.96E+01 | (1.38E+01) | (1036.31) | 2.96E-02 | (7.85E-03) | (72.94) |
| $w \to 0.7$ to $w \to 0.5$, $iw = sw = 2$ | 1.19E+01 | 2.30E+01 | 9978.35 | 0.00E+00 | 2.16E-02 | 6671.85 |
| | 3.28E+01 | (6.71E+00) | (96.82) | 6.82E-02 | (2.02E-02) | (1285.39) |
| Multi-swarm 1 (1, 2, 3) | 9.95E+00 | 2.47E+01 | 9627.25 | 0.00E+00 | 2.30E-02 | 5561.50 |
| | 7.56E+01 | (1.50E+01) | (729.14) | 6.10E-02 | (1.96E-02) | (691.97) |
| Multi-swarm 2 (2, 3, 4) | 1.59E+01 | 2.74E+01 | 8037.50 | 0.00E+00 | 3.21E-02 | 5083.35 |
| | 3.78E+01 | (6.96E+00) | (1767.09) | 8.04E-02 | (3.00E-02) | (366.67) |

Table 4. Summary of the most significant results obtained from the optimization of the 30-dimensional Schaffer f6 function using four selected parameters' settings and two multi-swarm optimizers (statistics out of 20 runs)

| OPTIMIZER | SCHAFFER F6 | | |
|---|---|---|---|
| | Best solution | Mean best solution | Mean time-steps |
| | Worst solution | (Standard deviation) | (Standard deviation) |
| $w = 0.7$, $iw = sw = 2$ | 7,82E-02 | 1,37E-01 | 9801,75 |
| | 1,78E-01 | (2,65E-02) | (521,13) |
| $w = 0.7298$, $iw = sw = 1.49609$ | 7,82E-02 | 2,07E-01 | 4288,60 |
| | 3,46E-01 | (7,20E-02) | (372,09) |
| $w = 0.5$, $iw = sw = 2$ | 3,72E-02 | 1,26E-01 | 4611,55 |
| | 1,78E-01 | (4,02E-02) | (516,55) |
| $w \to 0.7$ to $w \to 0.5$, $iw = sw = 2$ | 7,82E-02 | 1,17E-01 | 9161,60 |
| | 1,78E-01 | (3,05E-02) | (711,02) |
| Multi-swarm 1 (1, 2, 3) | 7,82E-02 | 1,50E-01 | 5887,15 |
| | 2,28E-01 | (4,15E-02) | (1752,03) |
| Multi-swarm 2 (2, 3, 4) | 7,82E-02 | 1,30E-01 | 7065,70 |
| | 1,78E-01 | (4,13E-02) | (2083,97) |

## 6. Conclusions

This paper analyzed the influence of the four parameters of the particles' velocity update equation of the basic particle swarm algorithm on the overall behaviour of the swarm. This is carried out taking into consideration the evolutions of the best solution found so far and of the average among the current population of candidate solutions; the particles' ability to—and speed of—clustering; and the optimizer's robustness in the sense of its reluctance to getting trapped in suboptimal solutions. Thus, four homogeneous optimizers with promising parameters' settings and two multi-swarm optimizers were proposed and tested on a set of benchmark functions.

For cases where the aim is to fine-tune the parameters of the basic PSO algorithm in order to carry out a refined search (in detriment of the optimizer's ability to escape sub-optimal solutions), the use of either the constriction factor or of the inertia weight $w < 1$—keeping the polynomial relationship stated in Eq.(9)—appears to be a good strategy. Notice that the inertia and the acceleration weights are correlated in both cases, and that the former is kept smaller than "1" in order to help control the explosion and to enhance the particles' fine-clustering.

Due to the random weights, it is recommended here that the acceleration weight is kept not greater than four. Greater values result in the increase of the influence of the random weights, and in the search becoming rather erratic instead of local, as it may seem at first glance. Besides, it appears convenient to keep the individuality and sociality weights equal to one another, letting the random weights alternate their relative importance. If a higher exploration at the beginning of the search and a progressive switch to a higher exploitation are sought, it is suggested here that this desired behaviour should be controlled by means of the inertia weight rather than by setting a higher individuality weight. A higher individuality might harm the exploration capability for some problems, for instance when the particles update their best experiences very frequently. Alternatively, a local version of the optimizer with small neighbourhoods which increase in size as the search progresses might be a good strategy in this regard.

For cases where the aim is to set the parameters of the basic PSO algorithm that lead to a system reluctant to get trapped in sub-optimal solutions, it seems appropriate to set the inertia weight so that the average behaviour of the swarm is somewhere between the cyclic behaviour and the one that favours fine-clustering (e.g. $1 < w < 0.5$ for $iw = sw = 2$).

Although the inertia weight and/or the constriction factor are able to either prevent the particles from exploding or to bring them back to the region of interest, it is advisable to restrict the maximum value that the components of the particles' velocities can take so as to avoid numerous unnecessary function evaluations. It is recommended to set a great value to the $v_{max}$ constraint so that it does not jeopardize the explorative ability of the swarm, affecting the normal trajectories of the particles as rarely as possible. A reasonable value can be $v_{max} = \frac{x_{max} - x_{min}}{2}$.

If the aim is to fine-tune the parameters for a general-purpose optimizer, the setting must result in the algorithm presenting both the ability to fine-cluster and to escape sub-optimal solutions, which do not typically come together. In this regard, the multi-swarm optimizers, composed of sub-swarms which present such abilities, seem to be an easy and effective strategy. This can be inferred from the results obtained when optimizing the benchmark functions in Table 1 (refer to Table 2 to Table 4). It is interesting to observe as well that the homogeneous optimizer with $iw = sw = 2$ and sigmoidly time-decreasing inertia weight from "0.7" (which favours exploration) to 0.5 (which favours fine-clustering)—as shown in Figure 12—also performs well on all the benchmark test functions considered.

**Acknowledgement**
The authors acknowledge the financial support from funding from the Objective 1 West Wales and the Valleys Programme from the European Social Fund through the Welsh European Funding Office.